# Bayesian Optimization for Materials Design with Mixed Quantitative and Qualitative Variables

**Yichi Zhang**
Mechanical Engineering
Northwestern University
Evanston, IL, US

**Daniel Apley**
Industrial Engineering and
Management Science
Northwestern University
Evanston, IL, US

**Wei Chen***
Corresponding Author
Mechanical Engineering
Northwestern University
Evanston, IL, US

**Abstract**

Although Bayesian Optimization (BO) has been employed for accelerating materials design in computational materials engineering, existing works are restricted to problems with quantitative variables. However, real designs of materials systems involve both qualitative and quantitative design variables representing material compositions, microstructure morphology, and processing conditions. For mixed-variable problems, existing Bayesian Optimization (BO) approaches represent qualitative factors by dummy variables first and then fit a standard Gaussian process (GP) model with numerical variables as the surrogate model. This approach is restrictive theoretically and fails to capture complex correlations between qualitative levels. We present in this paper the integration of a novel latent-variable (LV) approach for mixed-variable GP modeling with the BO framework for materials design. LVGP is a fundamentally different approach that maps qualitative design variables to underlying numerical LV in GP, which has strong physical justification. It provides flexible parameterization and representation of qualitative factors and shows superior modeling accuracy compared to the existing methods. We demonstrate our approach through testing with numerical examples and materials design examples. The chosen materials design examples represent two different scenarios, one on concurrent materials selection and microstructure optimization for optimizing the light absorption of a quasi-random solar cell, and another on combinatorial search of material constitutes for optimal Hybrid Organic-Inorganic Perovskite (HOIP) design. It is found that in all test examples the mapped LVs provide intuitive visualization and substantial insight into the nature and effects of the qualitative factors. Though materials designs are used as examples, the method presented is generic and can be utilized for other mixed variable design optimization problems that involve expensive physics-based simulations.

## Introduction

With advances in computational engineering, materials design and discovery have been increasingly viewed as optimization problems with the goal of achieving desired material properties or device performance[1–3]. One challenge of designing new materials systems is the co-existence of qualitative and quantitative design variables associated with material compositions, microstructure morphology, and processing conditions. While microstructure morphology can be described using quantitative, quantitative variables such as those associated with correlation function[4], descriptors[3,5,6], and spectral density functions[1,2], many composition and processing conditions are discrete and qualitative by nature. For example, in polymer nanocomposite design, there are numerous choices of material constituents (e.g., the types of filler and matrix) and processing conditions (e.g., the type of surface treatment); each combination follows drastically different physical mechanisms with significant impact on the overall properties[3,7]. As illustrated in Figure 1, the existence of both quantitative and qualitative material design variables results in multiple disjointed regions in the property/performance space. The combinatorial nature poses additional challenges in materials modeling and the search for optimal solution.

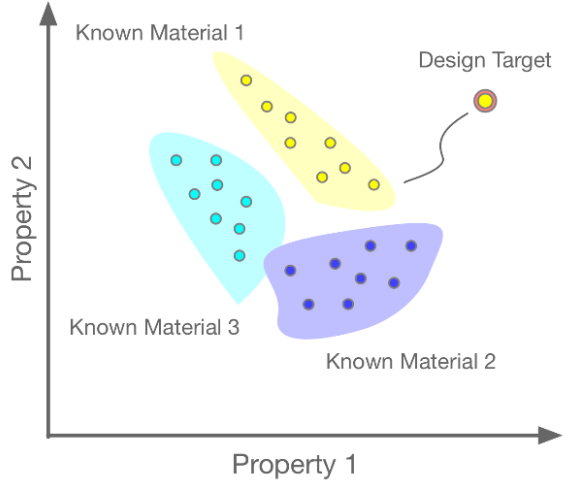

Figure 1: Target space exploration in materials design

In contrast to the traditional trial-and-error based experiment approach to materials design, computational materials design methods have emerged as an efficient and effective alternative in the past decade, building upon the advancement of simulation techniques such as finite element analysis (FEA) and density functional theory (DFT) that can accurately model and predict material properties at different length scales[8,9]. Recent years have seen new developments in computational methods for designing materials with quantitative and qualitative variables simultaneously, e.g., structure optimization and materials selection of a multilayer beam structure supported by a spring using a genetic algorithm (GA)[10], and optimization of a thermal insulation system using mixed variable programming (MVP) and a pattern search algorithm[11,12]. However, directly employing expensive simulation models in mixed-variable optimization is still restrictive because optimization algorithms commonly require hundreds of evaluations of the objective (material properties) for problems with only quantitative variables and the computational demand is significantly higher with mixed-variables.

To address the issue of expensive simulation, a common strategy in simulation-based optimization is to build a metamodel, a.k.a., response surface model, based on data generated by simulations, and then directly use the metamodel for optimization. Design of experiments (DOE) methods, such as Latin-hypercube sampling (LHS)[13], are often used to improve the overall metamodel accuracy by generating samples that cover the design space as evenly as possible. However, this is not the most efficient approach if the design objective is well defined because it is apparent that more sample points should be selected close to the "optimal" locations rather than uniformly over the whole design space.

In contrast, Bayesian Optimization (BO) provides an adaptive paradigm to sample the design space more efficiently for identifying the global optimum. In particular, a prior response surface model of the objective is prescribed and then sequentially refined as data are observed via an acquisition function[14]. One essential advantage of using BO for materials design is the emergence of various materials databases, such as the polymer nanocomposites data resource NanoMine[15,16], and the open quantum materials database (OQMD) that stores high-throughput DFT data[17]. These databases provide valuable low-cost existing knowledge as a "starting point" for Bayesian inference to guide the rapid exploration of novel material designs. Recent years have seen a number of extensions of using BO in materials design, such as the optimization of the synthesis process of short polymer fibers[18], adaptive optimization of the elastic modulus of the $MAX_2$ phase[19], and the prediction of crystal structures[20]. Nevertheless, these materials design applications of BO are all limited to considering only quantitative design variables, such as the constriction angle, channel width and solvent speed[18], the $s$-, $p$-, and $d$-orbital radii of atoms[19], and structure descriptors[20].

In real applications, most materials design scenarios involve qualitative or categorical factors, such as compositions (selection of material types) and particle surface treatment conditions (e.g., octyldimethylmethoxysilane and aminopropyldimethylethoxysilane) in nanodielectrics[7].

The challenge of including qualitative design factors within BO lies in response surface modeling in the absence of any direct ordering and distance metrics between the factor levels like the inherent distance metric for quantitative design variables. Gaussian process (GP) models, a.k.a., kriging models have become the most popular method for modeling simulation response surfaces[21–23] and thus are widely employed in BO frameworks, because of its flexibility to capture complex nonlinear response surface as well as to quantify uncertainties in prediction. However, these standard GP models are only applicable for quantitative inputs (i.e., quantitative design variables). Most state-of-the-art BO implementations use 0/1 dummy variables to represent qualitative input levels and then fit standard GP models with quantitative inputs, which is essentially equivalent to fitting a multi-response GP model where a different response surface (over the quantitative inputs) is assumed for each combination of qualitative factor levels. This has been shown to be restrictive theoretically and incapable of capturing complex correlations between qualitative levels when the number of levels is large[24–26].

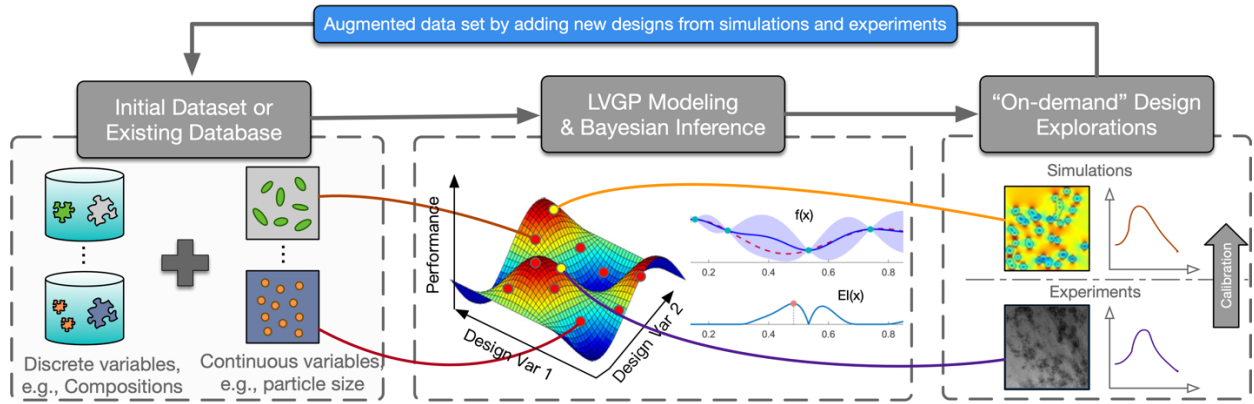

Figure 2: Bayesian optimization framework for data-driven materials design

To overcome the aforementioned limitations, we propose a BO framework for data-driven materials design with mixed qualitative and quantitative variables (Figure 2). The proposed framework is built upon a novel latent variable GP (LVGP) modeling approach that we recently developed for creating response surfaces with both qualitative and quantitative inputs[27]. The key idea of the LVGP approach is to map the qualitative factors into low-dimensional quantitative latent variable (LV) representations. Our study[28] showed that the LVGP modeling dramatically outperforms existing GP models with qualitative input factors in terms of predictive root mean squared error (RMSE). In addition to empirical evidence of far better RMSE predictive performance across a variety of examples, the LVGP approach has strong physical justification in that the effects of any qualitative factor on a quantitative response must *always* be due to some underlying quantitative physical input variables (otherwise, one cannot code the physics of the simulation model)[28]. Noting that the underlying physical variables may be extremely high-dimensional (which is why they are treated as a qualitative factor in the first place), the LVGP mapping serves as a low-dimensional LV surrogate for the high-dimensional physical variables that captures their collective effect on the response. In addition to outstanding predictive performance, this LV mapping provides an inherent ordering and structure for the levels of the qualitative factor(s), such as the type of material constituents and processing types, which can provide substantial insights into their influence on the material properties/performance. In this manner the LVGP approach can model a large number of qualitative levels with a relatively small number of parameters, which improves the prediction while maintaining low computational costs. Moreover, in contrast to the existing methods for handling qualitative factors[24–26,29], LVGP is compatible with any standard GP correlation functions for quantitative inputs, including nonseparable correlation functions such as power exponential, Matèrn and lifted Brownian.

In this study, we integrate the LVGP approach with a BO framework (we call it LVGP-BO) for data-driven materials design that involves both qualitative and quantative matereials design values. We examine the LVGP-BO approach over a variety of mathematical and real materials design examples and demonstrate its superior optimization performance over other state-of-the-art methods.

## Results

We present two mathematical examples and two materials design examples to demonstrate the efficacy of the proposed LVGP-BO approach for mixed-variable problems and the physical insights it brings. The proposed method is compared to the state-of-the-art BO implementation *bayesopt* in MATLAB[30], which represents qualitative factors as dummy variables first and then fits a standard GP model with quantitative inputs [26]. When comparing both methods, we start with the same set of initial datasets and stop at the maximal allowed number of iterations. The same procedure is repeated multiple times to assess the statistical stability of both methods. In both scenarios, expected improvement (EI)[31] is used as the acquisition function. We denote the LVGP-BO result as "LV-EI" and *bayesopt* as "MC-EI" because its GP model is equivalent to the multiplicative covariance (MC) model in literature[24,26].

### Materials Design Examples

*High-performance Light Absorbing Quasi-Random Solar Cell*

We first present a solar cell design problem, in which both the light scattering structure pattern and material selection are optimized simultaneously, in contrast to the original design problem that only considered tuning the scattering pattern represented by quantitative variables [2,32]. Figure 3 shows the setup of this solar cell design problem: the middle layer with the quasi-random structure is the light-trapping layer of thickness $t_1$, patterned on the amorphous silicon (a-Si) based absorbing layer with a total thickness of $t$. The bottom silver layer prevents the light escaping on the back side. The light-trapping structure is optimized for maximizing light absorption at 650nm by balancing the competing processes of light reflection and scattering. Besides designing the light trapping layer's pattern, we also consider the choices of materials. First, there are three types of a-Si with different refractive indices, and on top of the light trapping layer, is a 70 nm thin layer of anti-reflection coating (ARC), which can be chosen from three different materials. Table 1 lists the refractive indices (n) of the three a-Si's and three ARC's and the extinction coefficient (k) of a-Si. Many existing materials systems involve such concurrent materials selection and structure optimization decisions shown in this example.

Rigorous coupled wave analysis (RCWA) [33,34] is employed to evaluate the light absorption coefficients of the reconstructed structures. RCWA is a Fourier-domain-based algorithm that can solve the scattering problems for both periodic and aperiodic structures. The length of the unit cell for RCWA calculation is set at 2000 nm. The light trapping pattern is represented using the SDF (spectral density function) method as we proposed in [32], which significantly reduces the dimensionality of quasi-random microstructures. In this case, we use a uniform SDF defined by its left end $a$ and width $b$. These two parameters, along with the space filling ratio $\rho$, are the three design variables that describe the light trapping layer's pattern. The overall thickness $t$ of a-Si layer is fixed as 600nm, but the thickness of the light trapping layer $t_1$ is adjustable in this example. The feasible ranges of all design variables, including both quantitative and qualitative variables, are listed in Table 2.

Table 1: Refractive index for the candidate materials

| Type | a-Si | | ARC |
|---|---|---|---|
| | n | k | n |
| 1 | 3.3569 | 0.0168 | 2.7225 |
| 2 | 3.1891 | 0.0160 | 3.0625 |
| 3 | 3.0212 | 0.0151 | 3.4225 |
| 4 | 2.8534 | 0.0143 | 3.2400 |
| 5 | 2.6855 | 0.0134 | 3.0625 |

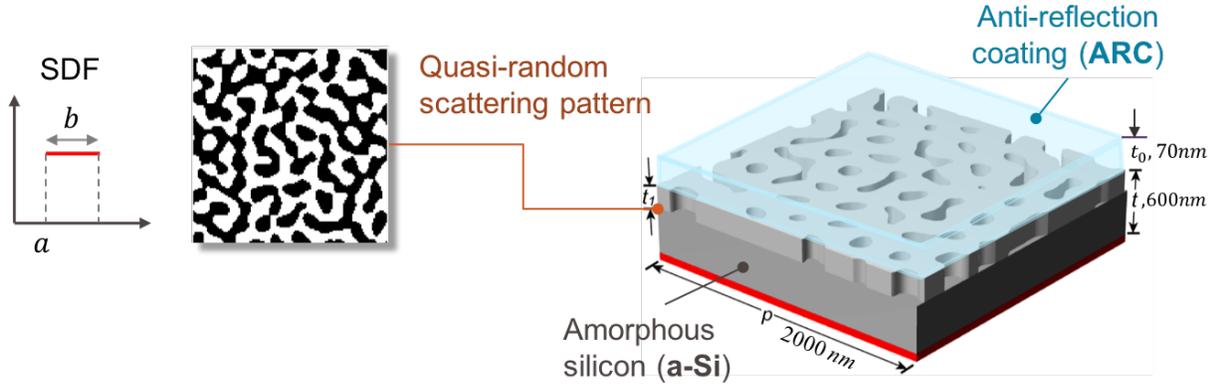

Figure 3: Setup of the solar cell design problem: the middle layer is the quasi-random scattering pattern with thickness $t_1$, directly etched on top of the a-Si substrate of a total thickness of t=600nm. The top layer in light blue is an ARC coating to reduce reflection of incident light. We use the SDF based approach to generate the quasi-random scattering pattern.

Table 2: Range of design variables in the solar cell design problem

| Quantitative Design Variable | Range |
| --- | --- |
| $t_1$ | $[50, 300]\,nm$ |
| $\rho$ | $[0.1, 0.9]$ |
| $a$ | $[0.0016, 0.0064]\,nm^{-1}$ |
| $b$ | $[0.0016, 0.0064]\,nm^{-1}$ |
| **Qualitative Design Variable** | **Levels** |
| a-Si Type | $\{1, 2, 3, 4, 5\}$ |
| ARC Type | $\{1, 2, 3, 4, 5\}$ |

In this design study, both the LV-EI and MC-EI methods start with 30 random initial samples where the quantitative variables $\{t_1, \rho, a, b\}$ are generated by the maximin LHD and the two materials type variables *a-Si Type* and *ARC type* are sampled uniformly. Both methods are terminated after the maximal allowed 100 iterations, and the same procedure is repeated for 20 replicates. Because there are certain uncertainties associated with the microstructure reconstruction process when generating the quasi-random scattering pattern using SDF, for each fixed pair of $\{\rho, a, b\}$, we generate three statistically equivalent microstructures and take the average of their light absorptions simulated in RCWA as the corresponding response. The acquisition function used in this case is EI with the "plug-in" $\mu_{min}(x)$, which is introduced in Method Section for the noisy response scenario. We use "EI" in our notation for simplicity. The left panel of Figure 4 displays the optimization convergence history of both methods over 100 iterations. It is apparent that the LV-EI method converged significantly faster and also consistently achieved better solutions compared to MC-EI. The best solution found possessing a light absorption coefficient of 0.94, with the optimized quantitative design variables $\{a^* = 0.0049 nm^{-1}, b^* = 0.0035 nm^{-1}, \rho^* = 0.875, t_1^* = 100.95 nm\}$ and the optimal choices of ARC and a-Si are type 5 and type 3 respectively. The right panel shows three random scattering structures using the optimal design variables.

The LVGP model represents the correlation between qualitative levels by the distances in the 2D latent variable space: larger distances means more difference and less correlation. In this test case, the estimated latent variables of ARC type and a-Si type are illustrated in Figure 5, which shows that five different ARC materials are positioned approximately in a straight line with the sequence 1-2-3-4-5, consistent with the differences in their refractive indices listed in Table 2. This provides some insight and indicates that the refractive indices are the dominating characteristics of the ARC design factor compared to other simulation inputs, in terms of its effect on the properties. The relationships between the five a-Si's are not straightforward at first glance, because n and k have coupled effects of the overall performance: the real part n quantifies refraction and the imaginary part k represents the loss of flux intensity in the medium, so higher n and lower k would result in better absorption. As shown in the LV plot, the 3$^{rd}$ type a-Si has very different effects on the response compared to the other four types, which makes sense because in Table 2 a-Si type 3 has both moderate n and k, while the others either have high n and low k or low n and high k values. The optimal solution also suggests that type 3 a-Si can lead to more absorption, which in turn, validates the estimated LV representation in Figure 5.

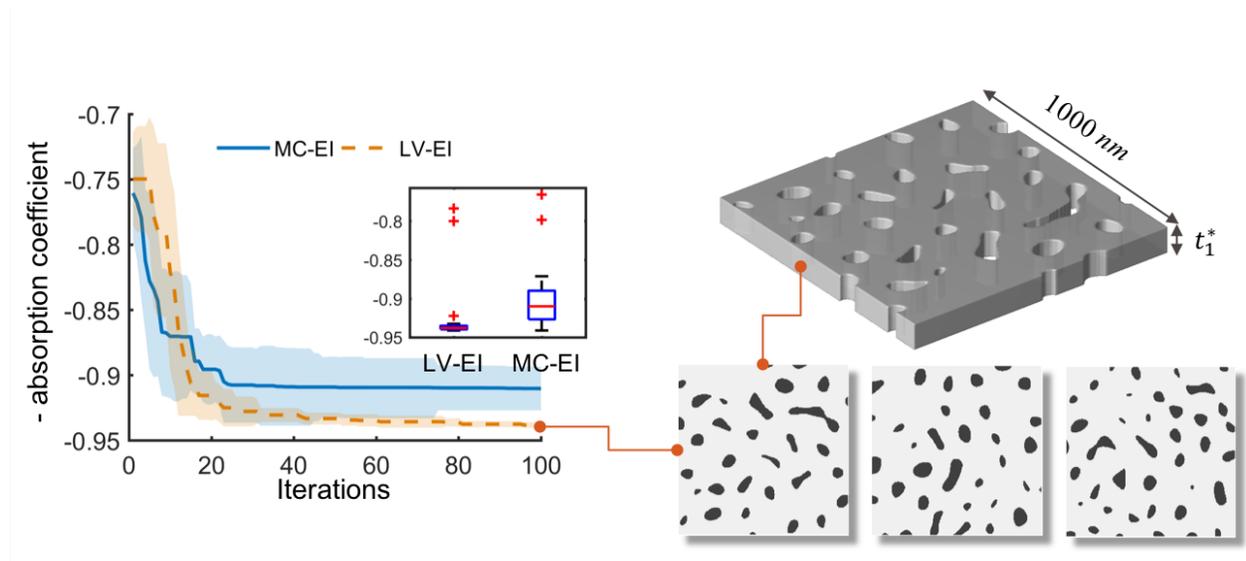

Figure 4: Optimization results of the light absorbing solar cell: the left panel displays the convergence plot of LV-EI and MC-EI method, with medians and median absolute deviations. The LV-EI outperforms MC-EI by consistently achieving better solutions with less uncertainty. The right panel shows three quasi-random patterns according to the results from LV-EI method. The top 3D structure is based on the first 2D pattern.

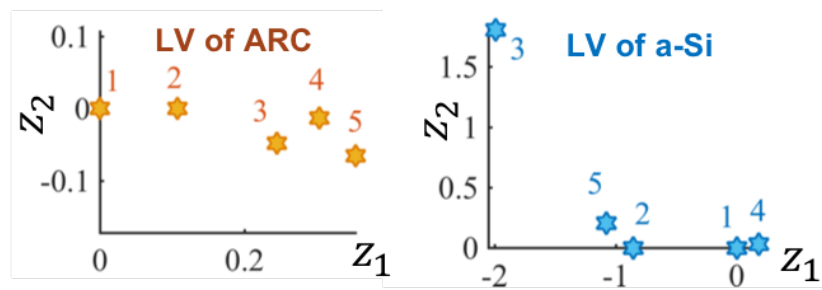

Figure 5: estimated latent variable of the two qualitative factors represent different materials: the LV of ARC approximately lines up in a straight line following the order 1-2-3-4-5, consistent with their refractive indices; from the LV of a-Si, type 3 is far from the other four, indicating a distinct influence on response compared to the others.

*Combinatorial Search of Hybrid Organic-Inorganic Perovskite (HOIP)*

We present here another materials design example adopted from literature with a combinatorial search of Hybrid Organic-Inorganic Perovskites (HOIPs)[35], where all design variables are qualitative. Such problems are quite popular in materials design with emphasis on design of materials constitutes. HOIPs are an exciting class of new materials that exhibit extremely promising photovoltaic (PV) properties. The goal is to search the perovskite compositional space for an $ABX_3$ combination with optimal intermolecular binding energy to a solvent molecule, $S_0$. The A-site cation has three candidates {MA, FA, Cs}, the B-site in HOIP is often occupied by metal cations and in this case, is fixed to be Pb. X denotes the halide, which has three options {Cl, Br, I} and in this design scenario, mixed halides are allowed, which means the three X's in $ABX_3$ can be different. There are eight different solvents $S_0$'s to explore, and the binding energies between $ABX_3$ and $S_0$ are results from DFT calculations.

There are five qualitative design variables: the first denotes the choice of the A-site cation (three levels), three other variables indicate the selection of each of the three halides in the configuration (three levels each), and the fifth variable represents the type of the solvent (eight levels). Out of all the possible combinations, 240 are stable, whose binding energies are precomputed by DFT. The code and dataset are made available by the authors of[35] at https://github.com/clancyLab/NCM2018, where the DFT calculations are done through the Physical Analytics Pipeline (PAL) developed by the Clancy Research Group from Cornell University[35]. Figure 6 (a) shows the distributions of the binding energies of the 240 samples, which indicates that majority of the samples have negative binding energies larger than -30, and the best solution has binding energy around -41.3. This ground truth was determined via exhaustive simulation, which the mixed-variable BO algorithms seeks to avoid. Their efficiency can be assessed based on whether the method can identify the true optimal solution from among the 240 combinations using as few evaluations as possible.

When comparing the LV-EI and MC-EI methods, we repeat our tests for 30 replicates, and in each replicate, both methods start with the same set of 10 initial random samples chosen from the samples with binding energy larger than -30 (to make the optimization more challenging), and the optimization was terminated after another 50 iterations. With respect to the capability of identifying the correct optimal solution, from Figure 6b, we note both methods work well in this combinatorial search problem: the objective function drops quickly within less than ten iterations, and in most of the replicates both methods found the global optimum. Nevertheless, the proposed LV-EI method outperforms MC-EI with smaller variance and quicker convergence. Moreover, the LV-EI method is more robust. Specifically, the boxplot in the inset shows that LV-EI found the exact best solution in 28 of the 30 replicates, whereas MC-EI found the exact best solution in only 22 of 30 replicates.

To better understand the correctness of the fitted LVGP model, we visualize the estimated LVs for solvent type in Figure 6d: solvent types 1 and 7 are located far from the other six types of solvents, which indicates they might have effects on the binding energy more distinct than the others. To validate this finding, we analyze the distribution for binding energies of the 240 samples in the dataset by singling out solvent types 1 and 7 in Figure 6c. All the samples with large binding energy ($> 30$) are formed with solvent types 1 and 7 and using other six types of solvents resulted in much smaller binding energies, which is consistent with our interpretation of the estimated LVs in Figure 6d. Our LVGP model successfully found that THTO is a superior solvent for dissolving ABX3. Using conventional approaches, it would be difficult to draw such design insights from the original raw dataset. In conclusion, the proposed mixed variable LVGP-based Bayesian optimization approach effectively searches a combinatorial design space, efficiently identifies the global optimal solution, and provides additional design insights through the LV representation of qualitative factors.

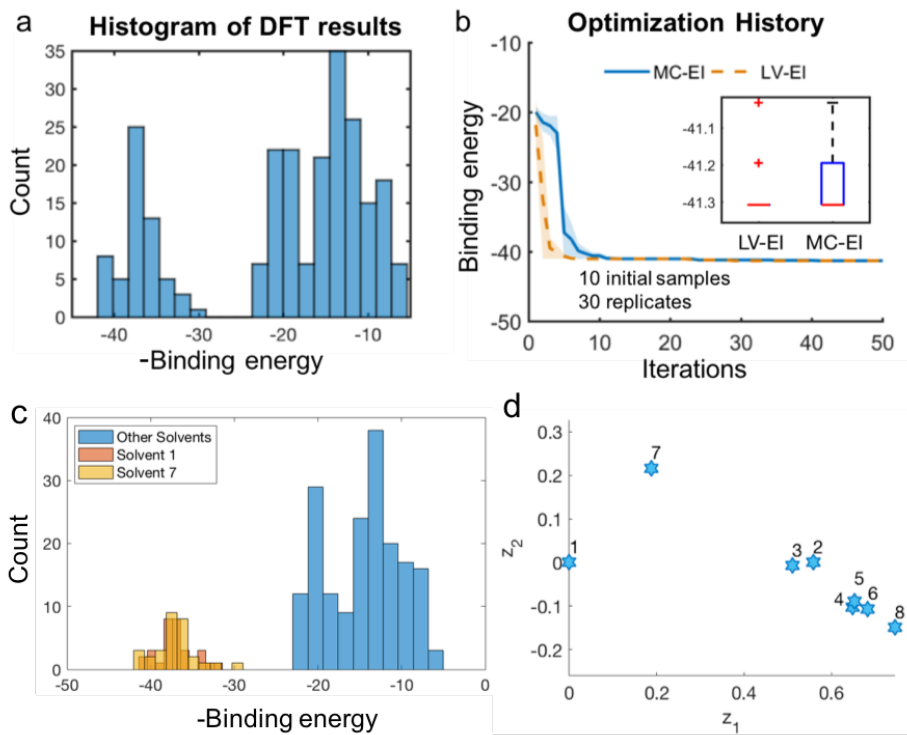

Figure 6: (a) histogram of the simulated binding energy from DFT, (b) optimization convergence plot of LV-EI and MC-EI methods with 10 initial random samples, median and median absolute deviation at each iteration are plotted, (c) negative binding energy distribution by solvent types, (d) estimated latent variable for solvent types.

## Mathematical Examples

Two mathematical examples are used to illustrate the capability of LVGP-BO in finding the global optimization solutions for mixed variable problems.

*Branin Function*

The first mathematical function considered is the Branin-Hoo function [36], which originally has two quantitative variables $x_1$ and $x_2$, while in this example we convert $x_2$ to be qualitative for testing purpose:

$$f(x_1,x_2) = \left(x_2 - \frac{5.1}{4\pi^2}x_1^2 + \frac{5}{\pi}x_1 - 6\right)^2 + 10\left(1 - \frac{1}{8\pi}\right)\cos(x_1) + 10,$$

where $x_1 \in [-5,10]$ and $x_2$ is qualitative, with four levels corresponding to the values $\{0, 5, 10, 15\}$. As plotted in Figure 7a this function has 6 local minimum and a global minimum $f(-2.6, 10) \approx 2.79118$. We can also observe that levels 1 and 2 of $x_2$ are closely correlated and levels 3 and 4 are closely correlated, while levels 1 and 3 are quite different from each other. Both methods start with 10 random initial points and continue sampling for another 30 iterations. Figure 7c shows the convergence history of the LV-EI and MC-EI method, from which we see that the LV-EI converges much faster than MC-EI and after 30 iterations LV-EI achieved a more accurate solution as illustrated in the inset of Figure 7.

*Gold-Stein Price Function*

The second mathematical example is the Goldstein-price function [36], which also has two input variables $x_1$ and $x_2$, where $x_2$ is made qualitative:

$$f(x_1,x_2) = \left[1 + (x_1 + x_2 + 1)^2 \left(19 - 14x_1 + 3x_1^2 - 14x_2 + 6x_1x_2 + 3x_2^2\right)\right],$$

where $x_1 \in [-2,2]$ and $x_2$ is qualitative, with five levels corresponding to the values $\{-2, 1, 0, 1, 2\}$. The global minimum is $f^*(0, -1) = 3$. In this example, both methods start with 20 random initial points, and continue sampling for another 30 iterations. Figure 7d shows the convergence history of the LV-EI and MC-EI method, from which, we see that LV-EI converges much faster than MC-EI consistently out of the 30 replicates. The LV-EI method is also more robust as it has a smaller variance than MC-EI. According to the inset boxplot, we also note that after 30 iterations, most of the 30 replicates of LV-EI are very close to the real global minimum 3 while MC-EI has a much larger gap to the real solution. The two numerical tests shown here illustrate our proposed LVGP based BO approach as an effective method for global optimization of mixed-variable problems.

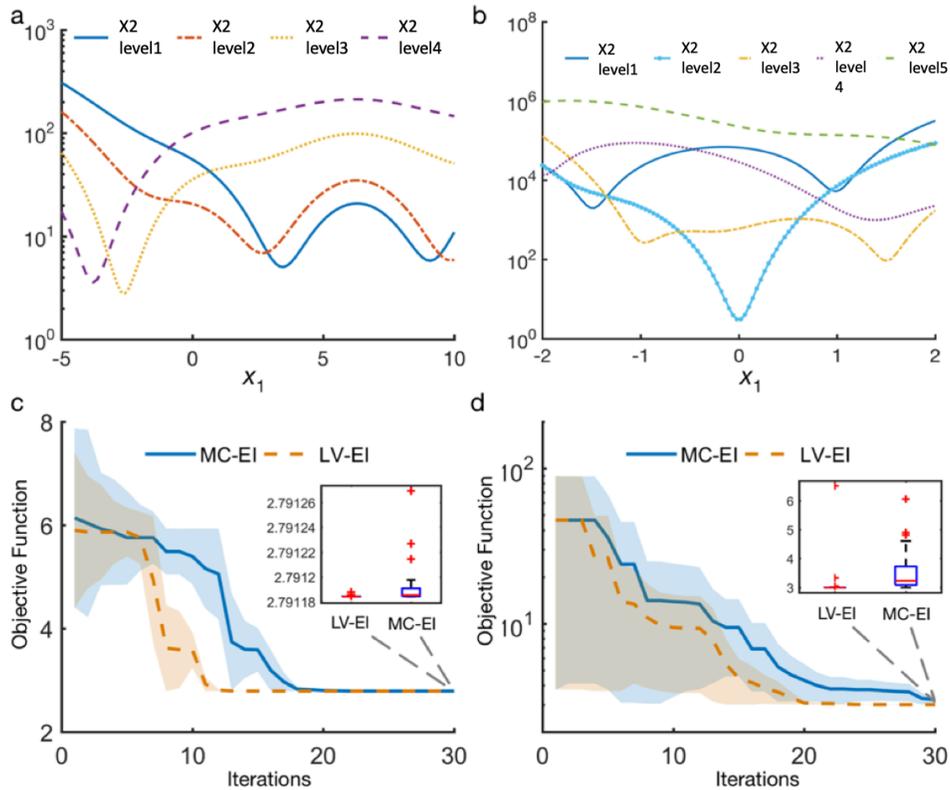

Figure 7: (a) plot of the Branin function, (b) plot of the Gold-Stein Price function, and problems having multiple local minima, and y-axes are in log-scale to visualize the local minimum better; (c) Optimization history of 30 replicates for the Branin example, starting with 10 initial sample points; (d) Optimization history of 30 replicates for the Gold-Stein Price example, with 20 initial sample points, y-axis in log scale. The middle lines represent the median values of the 30 replicates at each iteration, and the shaded bounds represent median +/- median absolute deviations (MAD). Insets in (c) and (d) display the boxplots of the minimum obtained after 30 iterations;

## Discussion

In this paper, we integrate a novel latent variable approach to GP modeling into Bayesian Optimization to support a variety of materials design applications with mixed qualitative and quantitative design variables. The LVGP approach not only provides superior model predictive accuracy compared to existing GP models, it also offers meaningful visualization of the correlation between qualitative levels (Figure 5 and Figure 6d). The proposed Bayesian Optimization approach is especially useful for materials design as it can fully utilize the existing materials databases and sequentially explores the unknown design space via Bayesian inference and "on-demand" simulations/experiments. Our proposed method overcomes the challenge of using BO for problems with qualitative input factors and has achieved superior performance compared to the state-of-the-art MATLAB BO implementation the using dummy variable representation for qualitative factors, which is demonstrated through both mathematical and real materials design examples (Figure 7c and d, Figure 4 and Figure 6b). For materials design, we successfully utilized the proposed BO framework to improve the light absorption of the quasi-random solar cell by designing the microstructure and selecting materials constituents simultaneously. Moreover, we showed that our method is efficient and effective for optimizing the material constituents of a hybrid organic-inorganic perovskite, a more challenging combinatorial search problem.

While this paper is focused on design of new materials and materials systems, the method presented is generic and can be used for other challenging engineering optimization problems where qualitative and quantitative design variables co-exist. The current BO framework will be further extended for multi-objective problems and the inclusion of physical constraints by refining the sampling strategy in association with the acquisition function.

## Methods

### Latent Variable Gaussian Process (LVGP) for Both Qualitative and Quantitative Factors

We first briefly review the technical details of the standard GP model for quantitative variables, and then describe our novel LVGP model to handle qualitative factors. To make the discussion more concrete, let $y(\cdot)$ denote the true physical response surface model with inputs $\boldsymbol{w} = (\boldsymbol{x}, \boldsymbol{t})$ where $\boldsymbol{x} = (x_1, \ldots, x_p) \in \mathbb{R}^p$ represents $p$ quantitative variables and qualitative factors $\boldsymbol{t} = (t_1, \ldots, t_q) \in \{1, 2, \ldots, m_1\} \times \{1, 2, \ldots, m_2\} \times \ldots \times \{1, 2, \ldots, m_q\}$, where the $j$th qualitative factor $t_j$ has $m_j$ levels that are coded (without loss of generality) as $\{1, 2, \ldots, m_j\}$. A GP model with only quantitative input variables is commonly assumed to be of the form:

$$y(\boldsymbol{x}) = \mu + Z(\boldsymbol{x}), \qquad (1)$$

where $\mu$ is a constant prior mean term, $Z(\cdot)$ is a zero-mean Gaussian process with stationary covariance function $K(\cdot,\cdot) = \sigma^2 R(\cdot,\cdot)$, $\sigma^2$ is the prior variance, and $R(\cdot,\cdot) = R(\cdot,\cdot \mid \boldsymbol{\phi})$ denotes the correlation function with parameters $\boldsymbol{\phi}$. A Gaussian correlation function is commonly used:

$$R(\boldsymbol{x}, \boldsymbol{x}') = \exp\left\{-\sum_{i=1}^{p} \phi_i (x_i - x_i')^2\right\}, \qquad (2)$$

which represents the correlation between $Z(\boldsymbol{x})$ and $Z(\boldsymbol{x}')$ for any two input locations $\boldsymbol{x} = (x_1, \ldots, x_p)$ and $\boldsymbol{x}' = (x_1', \ldots, x_p')$, where $\boldsymbol{\phi} = (\phi_1, \ldots, \phi_p)^T$ is the vector of correlation parameters to be estimated via MLE, along with $\mu$ and $\sigma^2$. The correlation between $y(\boldsymbol{x})$ and $y(\boldsymbol{x}')$ depend on the spatial distance between $\boldsymbol{x}$ and $\boldsymbol{x}'$ and the correlation parameters. Other choices of correlation functions include power exponential, Matèrn [37] and lifted Brownian [38]. These types of correlation functions cannot be directly applied to problems with qualitative factors because the distances between levels of qualitative factors are not defined, and the levels have no natural ordering since $t$ are assumed to be nominal categorical factors.

The LVGP approach[28] provides a natural and convenient way to handle qualitative input variables by mapping the levels of each qualitative factor $t$ to a 2-dimensional (2D) continuous latent space. This has strong physical justification, which may explain the outstanding predictive performance of the LVGP approach[28]: For any real physical system with a qualitative input factor $t$, there are always underlying quantitative physical variables $\{v_1, v_2, ...\} = \{v_1(t), v_2(t), ...\}$ (perhaps very high-dimensional, poorly understood, and difficult to treat individually as quantitative input variables) that account for the differences between the responses at different levels of the qualitative factors. As shown in Figure 8, the three qualitative levels are associated with points in a high-dimensional space of $\{v_1, v_2, ...\}$, and the distances between the three points indicates the differences between the three levels. Our LVGP model uses a low-dimensional (2D) representation $\mathbf{z}(t) = g(v_1(t), v_2(t), ...)$ to approximate the actual distances between the three levels, which implicitly makes the rather mild assumption that collective effects $\{v_1(t), v_2(t), ...\}$ on $y$, as $t$ varies across its levels, can be captured by some low-dimensional function $\mathbf{z}(t) = g(v_1(t), v_2(t), ...)$. In many applications, a two-dimensional representation suffices to approximate the high dimensional data [39].

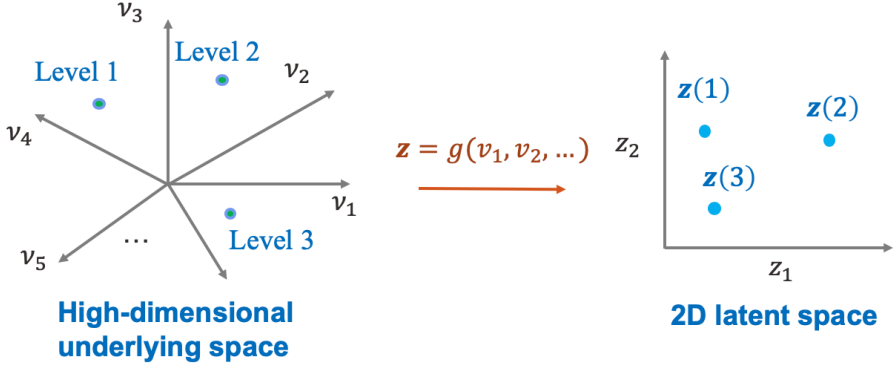

Figure 8: For a single factor $t$ with three levels, depiction of the mapping from the true high-dimensional underlying quantitative variables to the 2D latent variables

When there are multiple qualitative factors, let $\mathbf{z}^j(t_j) = (z_1^j(t_j), z_2^j(t_j))$ denotes the 2D mapped LV for the qualitative factor $t_j$. For a Gaussian correlation function, the LVGP approach then assumes the correlation function

$$cor\left(y\left(\mathbf{x}, \mathbf{t} = (t_1, \ldots, t_q)\right), y(\mathbf{x}', \mathbf{t}' = t_1', \ldots, t_q')\right)$$
$$= \exp\left\{-\sum_{i=1}^{p} \phi_i(x_i - x_i')^2 - \sum_{j=1}^{q} \left\|\mathbf{z}^j(t_j) - \mathbf{z}^j(t_j')\right\|_2^2\right\}, \quad (3)$$

where $\|\cdot\|$ denotes the $L_2$ norm. Note that the 2D mapped LVs $\mathbf{z}^j(t_j) = (z_1^j(t_j), z_2^j(t_j))$ are unknown and will be estimated along with $\mu$, $\sigma^2$ and $\boldsymbol{\phi}$ through maximum likelihood estimation (MLE). Under model (3), the log-likelihood function is:

$$l(\mu, \sigma^2, \boldsymbol{\phi}, \mathbf{Z}) = -\frac{n}{2}\ln(2\pi\sigma^2) - \frac{1}{2}\ln|\mathbf{R}(\boldsymbol{\phi}, \mathbf{Z})|$$
$$-\frac{1}{2\sigma^2}(\mathbf{y} - \mu\mathbf{1})^T \mathbf{R}(\boldsymbol{\phi}, \mathbf{Z})^{-1}(\mathbf{y} - \mu\mathbf{1}), \quad (4)$$

where $n$ is the sample size, $\mathbf{1}$ is an $n$-by-1 vector of ones, $\mathbf{y}$ is the $n$-by-1 vector of observed response values, $\mathbf{Z} = (\mathbf{Z}^1, \ldots, \mathbf{Z}^q)$, $\mathbf{Z}^j = \left(\mathbf{z}^j(1), \ldots, \mathbf{z}^j(m_j)\right)$ represents the values of the LV corresponding to the $m_j$ levels of the qualitative variable $t_j$, and $\mathbf{R} = \mathbf{R}(\boldsymbol{\phi}, \mathbf{Z})$ is the $n$-by-$n$ correlation matrix whose elements are obtained by plugging pairs of the $n$ sample values of $(\mathbf{x}, t)$ into (3). The MLE of $\mu$ and $\sigma^2$ in (4) can be represented in terms of the correlation matrix :

$$\hat{\mu} = (\mathbf{1}^T \mathbf{R}^{-1} \mathbf{1})^{-1} \mathbf{1}^T \mathbf{R}^{-1} \mathbf{y}$$
$$\hat{\sigma}^2 = \frac{1}{n}(\mathbf{y} - \hat{\mu}\mathbf{1})^T \mathbf{R}^{-1}(\mathbf{y} - \hat{\mu}\mathbf{1}). \tag{5}$$

Substituting the above into (4) and neglecting constants, the log-likelihood function becomes:

$$l(\boldsymbol{\phi}, \mathbf{Z}) \sim -n\ln(\hat{\sigma}^2) - \ln|\mathbf{R}(\boldsymbol{\phi}, \mathbf{Z})|, \tag{6}$$

which is maximized over the correlation matrix $\mathbf{R}$ that depends on the correlation parameters $\boldsymbol{\phi}$ and the values of the mapped latent variables in $\mathbf{Z}$.

After the MLEs of $\boldsymbol{\phi}$ and $\mathbf{Z}$ are obtained, the LVGP response predictions at any new point $\mathbf{x}^*$ is

$$\hat{y}(\mathbf{x}^*) = \hat{\mu} + \mathbf{r}(\mathbf{x}^*)\mathbf{R}^{-1}(\mathbf{y} - \hat{\mu}\mathbf{1}), \tag{7}$$

where $\mathbf{r}(\mathbf{x}^*) = \left(R(\mathbf{x}^*, \mathbf{x}^{(1)}), \dots, R(\mathbf{x}^*, \mathbf{x}^{(n)})\right)$ is a vector of the pairwise correlation between $\mathbf{x}^*$ and each data point $\mathbf{x}^{(j)}, j = 1, \dots, n$. Moreover, to quantify predictive uncertainty, the variance of the error for this prediction is:

$$\hat{s}^2(\mathbf{x}^*) = \hat{\sigma}^2 (r(\mathbf{x}^*) - \mathbf{r}(\mathbf{x}^*)\mathbf{R}^{-1}\mathbf{r}(\mathbf{x}^*)^T). \tag{8}$$

When the actual model is non-deterministic, we add an extra "nugget" parameter $\lambda$ to each diagonal element of the correlation matrix $\mathbf{R}$ to account for the noise of the response and it is estimated along with $\boldsymbol{\phi}$ and $\mathbf{Z}$.

## LVGP-BO Framework for Materials Design

Our proposed Bayesian Optimization framework for data-driven materials design consists four major steps (as shown in Figure 1): (1) Step 1 involves creating a materials dataset (either physical or/and computer data) based on the information gathered from literature, lab experiments and simulations, (2) Step 2 fits the LVGP model using the available dataset and provides uncertainty quantification of model prediction based on the nature of data, (3) Step 3 makes inference about where to sample the next point (either physical or computer experiments) based on an acquisition function that balances sampling where the response appears to be optimized (exploitation) vs. where the predictive uncertainty is high (exploration), and (4) Step 4 evaluates the chosen design point(s) to augment the materials database and update the metamodel. As this procedure keeps going, more sample points will be sequentially added to update the metamodel prediction and identify the global optimum solution.

There are two key components of the BO framework, a metamodel that provides predictions with uncertainty quantification, and a criterion that determines where to sample next. The LVGP model introduced in the previous section provides robust approximations of the actual response surface model with mixed variable types, as well as the uncertainty quantification. The next critical step is to consider where to sample next based on inferences of the fitted model (Step 3 "on-demand" design exploration on Figure 2), through a measure of the value of the information gained from sampling at a certain point, known as acquisition functions. Three commonly used acquisition functions in the literature are expected improvement (EI)[31], probability of improvement (PI), and lower confidence bound (LCB). For deterministic responses, expected improvement (EI) is the most widely used and works well over a variety of problems, while for noisy responses, EI with plug-in and knowledge gradient (KG) are proper choices [36,40]. Detailed benchmark studies of different acquisition functions are available in [36].

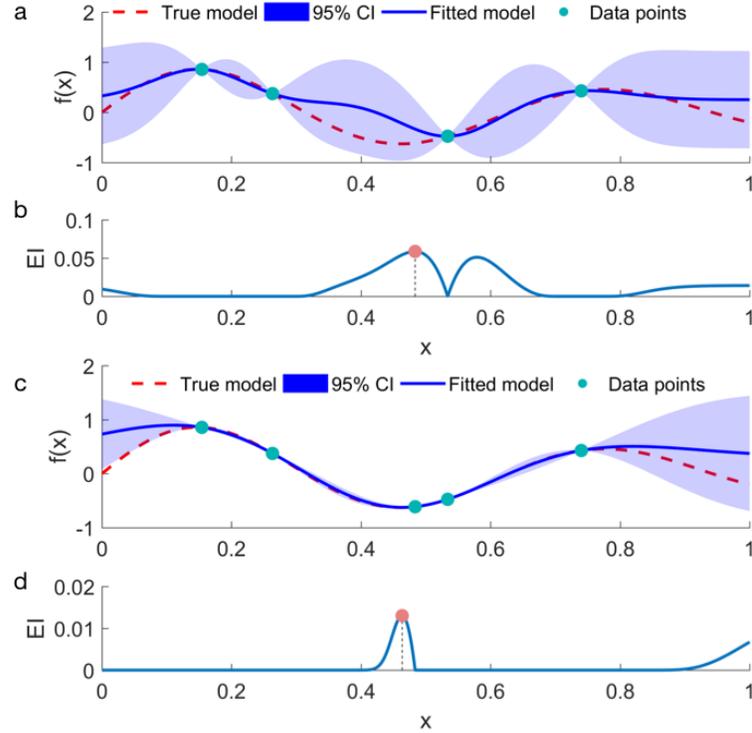

Figure 9: GP model evolution using EI as the acquisition function: (a) initial GP model based on four data points, (b) EI of the initial GP model, (c) updated GP model after sampling one additional data point based on EI, (d) EI of the updated GP model

EI balances exploitation and exploration as elaborated in Figure 9: a GP model is first fitted based on the four sample points (Figure 9a), and the fitted mean prediction $\hat{y}(x)$ is the blue curve, while the actual response $y(x)$ is the red dashed curve. Instead of picking the minimizer of $\hat{y}(x)$ as the next sample point, EI also considers the model uncertainties. Mathematically, EI quantifies the possible improvement of a particular point by incorporating both $\hat{y}(x)$ and the associated uncertainty $\hat{s}^2(x)$:

$$EI(x) = E[\max(0, \Delta(x))] = \hat{s}(x)\phi\left(\frac{\Delta(x)}{\hat{s}(x)}\right) + \Delta(x)\Phi(\frac{\Delta(x)}{\hat{s}(x)}), \qquad (10)$$

where $\hat{y}(x)$ and $y_{\min}$ are the mean prediction of the fitted GP model and the minimal value observed so far, $\Delta(x) = y_{\min} - \hat{y}(x)$, $\phi$ and $\Phi$ are the probability density function (PDF) and cumulative density function (CDF) of standard normal distribution. The maximizer of EI in Figure 9b is chosen as the next sampling point and a new GP model is fitted with this additional data point (shown in Figure 9c). The updated mode provides a more accurate approximation of the true model with much less uncertainty. The updated EI profile (Figure 9d) also indicates that the region around the true response optimizer has a large expected improvement. This BO algorithm is summarized in Table 3.

Table 3: Bayesian Optimization Algorithm

---

(0) Generate initial dataset $\boldsymbol{D_0}$
(1) **For** $n = 1, 2, \ldots,$ **do**
(2)     Fit the latent variable GP model $\hat{\boldsymbol{y}}_n(\boldsymbol{x}; \boldsymbol{D_{n-1}})$
(3)     Select the next sampling point $\boldsymbol{x_{n+1}}$ by maximizing *EI*:
$$\boldsymbol{x_{n+1}} = \arg\max_{x} EI(\boldsymbol{x}; \hat{\boldsymbol{y}}_n)$$
(4)     Query simulations/experiments to obtain $\boldsymbol{y_{n+1}}$
(5)     Augment data $\boldsymbol{D_{n+1}} = \{\boldsymbol{D_n}, (\boldsymbol{x_{n+1}}, \boldsymbol{y_{n+1}})\}$
(6) **End for**.

---

## Acknowledgements

This work was supported in part by National Science Foundation (NSF) Grants including CMMI 1662435, EEC 1530734, and DMREF 1729743.

## Author Contributions

The LVGP approach was developed jointly by Y. Zhang, D. Apley, and W. Chen. Y. Zhang and W. Chen formulated the idea of extending this approach to Bayesian Optimization. Code implementation was done by Y. Zhang, and the applications to numerical examples and material design problems were completed by Y. Zhang with continuous discussions and iterations with W. Chen and D. Apley. All authors reviewed the manuscript and contributed to the revision.

## Competing financial interests

The authors declare no competing financial interests.

## Data availability statement

The source code of this work will be made available upon request to the corresponding author.